\documentclass[conference]{IEEEtran}
\usepackage{lipsum} %
\IEEEoverridecommandlockouts
\usepackage{cite}
\usepackage{amsmath,amssymb,amsfonts}
\usepackage{algorithmic}
\usepackage{graphicx}
\usepackage{textcomp}
\usepackage{xcolor}
\usepackage{url}
\usepackage{hyperref}
\usepackage{booktabs}
\def\BibTeX{{\rm B\kern-.05em{\sc i\kern-.025em b}\kern-.08em
    T\kern-.1667em\lower.7ex\hbox{E}\kern-.125emX}}

\newcommand{\ours}{\textsc{CoT-Sep}}

\usepackage{xcolor}
\definecolor{pastelblue}{RGB}{173, 216, 230} 
\newcommand{\eg}{\textit{e.g., }}
\newcommand{\ie}{\textit{i.e., }}

\begin{document}

\title{Can Separators Improve Chain-of-Thought Prompting?\\
}

\author{
\IEEEauthorblockN{
Yoonjeong Park\IEEEauthorrefmark{1}\textsuperscript{,\textsection}, 
Hyunjin Kim\IEEEauthorrefmark{1}\textsuperscript{,\textsection}, 
Chanyeol Choi\IEEEauthorrefmark{2}, 
Junseong Kim\IEEEauthorrefmark{2}\textsuperscript{,\P{}}, 
Jy-yong Sohn\IEEEauthorrefmark{1}\textsuperscript{,\P{}}
}
\\
\IEEEauthorblockA{
\IEEEauthorrefmark{1}\textit{Yonsei University, Seoul, Republic of Korea}\\ 
\{dbw2140, hjhyunjinkim, jysohn1108\}@yonsei.ac.kr
}
\IEEEauthorblockA{
\IEEEauthorrefmark{2}\textit{Linq, Cambridge, Massachusetts, United States} \\
\{jacob.choi, junseong.kim\}@getlinq.com
}
}
\maketitle
\begingroup\renewcommand\thefootnote{\textsection}
\footnotetext{Equal contribution\\
\indent\textsuperscript{\P{}}Co-corresponding authors}
\endgroup

\begin{abstract}
Chain-of-thought (CoT) prompting is a simple and effective method for improving the reasoning capabilities of 
Large Language Models (LLMs). 
The basic idea of CoT is to let LLMs break down their thought processes step-by-step by putting exemplars in the input prompt. 
However, the densely structured prompt exemplars of CoT may cause the cognitive overload of LLMs. Inspired by human cognition, we introduce \ours{}, a method that strategically employs separators at the end of each exemplar in CoT prompting. These separators are designed to help the LLMs understand their thought processes better while reasoning. Interestingly, it turns out that \ours{} significantly improves the LLMs' performances on complex reasoning tasks  (e.g., GSM8K, AQuA, CSQA), compared with the vanilla CoT, which does not use separators. We also study the effects of the type and the location of separators tested on multiple LLMs, including GPT-3.5-Turbo, GPT-4, and LLaMA-2 7B. Interestingly, the type/location of separators should be chosen \textit{appropriately} to boost the reasoning capability of CoT.
The code is available at \href{https://github.com/cottonlove/CoT-SEP/}{\texttt{https://github.com/cottonlove/CoT-SEP}}
\end{abstract}


\section{Introduction}
\begin{figure*}[ht]
\begin{center}
\includegraphics[scale=0.32]{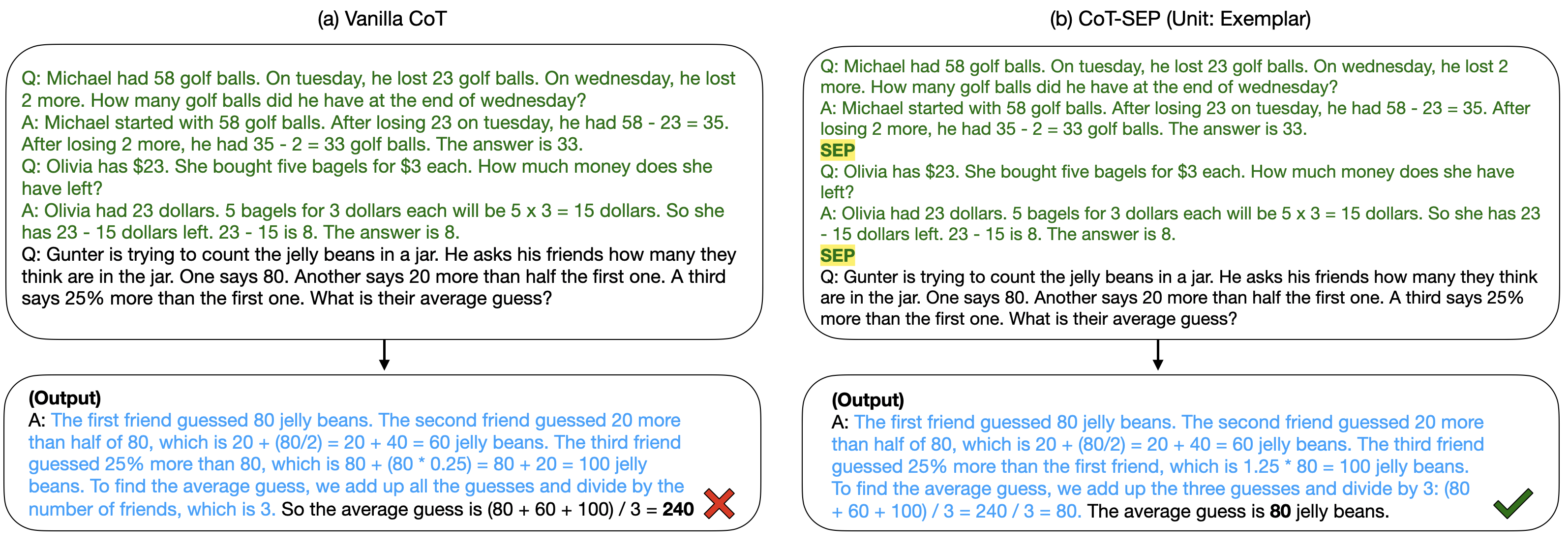} 
\end{center}
\caption{Comparison of the Vanilla CoT~\cite{wei2022chain} and our \ours{} for a test data sampled from the GSM8K benchmark \cite{cobbe2021training}, where two exemplars (colored as green texts) are provided in the input prompt. For \ours{}, a separator (denoted as SEP, usually chosen as `\textbackslash{n}' or 
 `\texttt{<br>}') is added at the end of each exemplar for better readability. Interestingly, adding the separators is enough to fix the issue of CoT.
}
\label{fig:figure1}
\end{figure*}

\noindent The use of large language models (LLMs) has significantly transformed our methods of processing information, enhancing performance across various fields. A key development in this area is Chain-of-Thought (CoT) prompting \cite{wei2022chain}, which enables LLMs to process complex reasoning. By outlining thought processes in a step-by-step manner, CoT prompting enables LLMs to demonstrate more sophisticated reasoning. By reasoning step-by-step, similar to the human cognition process, LLMs are able to tackle complicated problems with enhanced precision. \\
\\
A noteworthy observation in the current implementation of CoT prompting is the densely structured few-shot exemplars within a single prompt (see the left column of Fig.~\ref{fig:figure1}). While this approach gives a broad context to the LLMs, it may also cause a cognitive overload, ultimately limiting the LLM's capability to process and analyze information efficiently. \\
\\
In human cognition, the memory span is a fixed number of chunks, and therefore is important to organize the input sequence into chunks \cite{miller1956magical}. Thus, chunking through strategic separations and breaks in text, such as a linebreak, plays an essential role in human comprehension and enhancing reasoning. Inspired by this psychological concept, we introduce \ours{}, an approach that strategically inserts separators at the optimal positions. \ours{} puts separators at the position where the LLM can segment information into manageable portions, therefore enabling better comprehension by the LLM. \ours{} leads to a 
improvement in the performance of LLMs over CoT prompting, underscoring the importance of structured formatting for optimizing LLM outputs. We conduct evaluations on multiple separators on arithmetic reasoning benchmarks and a commonsense reasoning benchmark, 
revealing that inserting separators at an adequate location and creating a pattern for LLMs to decipher outperforms the original CoT prompting technique. \\
\\
Various prompting methods are proposed to improve the performance of CoT and its variants~\cite{ling2023deductive,long2023llmtot, besta2023got,weng2023large,zhang2023selfconvinced}, which introduce additional modules for iterative refinement and verification of intermediate thoughts. Another line of works proposes refining input questions \cite{xi2023selfpolish,deng2023rephrase} to let LLMs better interpret the target reasoning tasks. However, these existing methods require multiple iterations through LLMs, leading to increased expenses and extended timeframes. Additionally, their task-specific designs make practical applications challenging. 
Through extensive experiments, we confirmed that \ours{} is straightforward and effective in improving CoT prompting, by simply introducing a separator (SEP) between exemplars.\\
\\
Our empirical results on \ours{} indicate that using separators between exemplars in CoT prompting significantly aids the reasoning process of LLMs. For example, adding separators increases the accuracy by $2.8\%$ on AQuA dataset and by $1.3 \%$ on GSM8K dataset, when tested on GPT-3.5-Turbo model. 
Interestingly, we discover that the proper placement of separators in prompts considerably impacts the LLM's reasoning capabilities, potentially setting a new standard in the design of prompts for LLMs.

\section{Related Work}
\noindent \textbf{Large Language Model (LLM) Reasoning}
Complex tasks involving logical thinking, particularly in solving mathematical problems, are commonly challenging for natural language processing (NLP) models. \cite{cobbe2021training, koncel-kedziorski-etal-2016-mawps, patel-etal-2021-nlp}. Recent progress in the development of LLMs \cite{touvron2023llama, zhang2022opt, chowdhery2022palm, wei2022emergent, brown2020language} demonstrates remarkable capabilities in complex reasoning tasks. These works suggest that instead of directly generating final answers, LLMs perform better on reasoning tasks when guided through a step-by-step process. This approach involves using examples in few-shot settings and employing effective prompting strategies, such as the Chain-of-Thought (CoT) \cite{wei2023chainofthought} prompting. To enhance the step-by-step reasoning process further, recent studies have employed external modules \cite{bostrom2022natural, creswell2022faithful, tafjord2021proofwriter, lyu2023faithful, chen2023program} 
to verify and refine intermediate thoughts. Similar to these previous work, our work uses the step-by-step reasoning process and focuses on aiding the process for enhanced reasoning. \\
\\
\textbf{In-Context Learning} 
Large Language Models (LLMs) have demonstrated impressive abilities in few-shot learning \cite{lu-etal-2023-makes, qiao-etal-2023-reasoning}. In particular, by incorporating a few exemplars into the prompts, LLMs can show exceptional performance without the need for finetuning on a training dataset \cite{weng2023large2, wang-etal-2023-towards}, and such LLMs' ability is termed as in-context learning (ICL). However, these approaches face challenges when encountering tasks that demand complex reasoning. This led to the emergence of CoT \cite{wei2022chain}, which is an approach that generates a series of reasoning steps to arrive at the final answer and is highly effective in solving difficult tasks. Recent work has also highlighted the in-context abilities of LLMs when combined with CoT prompting \cite{wang2023selfconsistency, ling2023deductive}. Building upon in-context learning, we use separators in exemplars to enhance complex reasoning.

\section{\ours{}}

\noindent Consider one's ability to process complex information. It is common to break down the information into chunks for ease of processing \cite{miller1956magical}, and using separators can significantly enhance one's comprehension and readability by providing visual breaks. Inspired by this observation, we introduce \ours{}, a method that places separators at the end of each prompt exemplar. Similar to Chain-of-Thought prompting, \ours{} generates a logical sequence of intermediate reasoning steps leading to the final answer. In contrast, as shown on the right of Fig. \ref{fig:figure1}, when we include examples of chain-of-thought sequences into the prompt exemplars, we place separators at the end of each exemplar, which enables the LLMs to process a large amount of information in that manner.

\section{Experiments}

\subsection{Experimental Settings}
\noindent Our experiments are heavily based on the Vanilla CoT prompting \cite{wei2022chain}, and we use OpenAI's \verb|gpt-3.5-turbo-0613|\footnote{\label{openai}https://platform.openai.com/docs/models} for the results in Table \ref{table1} and \ref{table:ablation}. For the results in Table~\ref{table2}, we use Meta's \verb|llama-2-7b|\footnote{https://llama.meta.com/llama2}, OpenAI's \verb|gpt-4-0613|\textsuperscript{\ref{openai}} and \verb|gpt-4-turbo|, specifically the version of \verb|gpt-4-0125-preview|\textsuperscript{\ref{openai}}. For our experiments, we use benchmarks where CoT prompting leads to substantial improvement over standard prompting \cite{wei2022chain}.
To be specific, we test on two challenging mathematical reasoning task benchmarks (GSM8K~\cite{cobbe2021training} with 1319 samples and AQuA~\cite{ling2017program} with 254 samples) and
one commonsense reasoning benchmark (CSQA~\cite{talmor-etal-2019-commonsenseqa} with 1221 samples). 
We follow the prompts designed by \cite{wei2022chain} and added separators to test \ours{}.
More details regarding the exemplars used in our experiments are included in \url{https://github.com/cottonlove/CoT-SEP}.
\begin{table*}[t]
\caption{Accuracies ($\%$) of vanilla CoT, \ours{}, and Heterogeneous \ours{}, test on GPT-3.5-Turbo.
    For each task, \ours{} outperforms vanilla CoT. Here, the type of separator is written in the parenthesis, and Heterogeneous \ours{} uses different separators for different exemplars.
    }
    \centering
    \small
    \vspace{0.4cm}
    \label{experimental-result}
    \begin{tabular}{@{}ccccc@{}}
        \toprule
        & & \multicolumn{2}{c}{Arithmetic Reasoning} & \multicolumn{1}{c}{Commonsense Reasoning} \\ \cmidrule(r){3-4} \cmidrule(r){5-5}
        Method & & \multicolumn{1}{c}{GSM8K} & \multicolumn{1}{c}{AQuA} & \multicolumn{1}{c}{CSQA} \\ 
        \midrule 
        \multicolumn{2}{l}{Vanilla CoT~\cite{wei2022chain}} & 70.4 {\scriptsize $\pm$0.17} & 46.5 {\scriptsize $\pm$0.82} & 76.5 {\scriptsize $\pm$0.14} \\
        \multicolumn{2}{l}{\ours{} (\texttt{TripleSkip}, \ie  \textbackslash{n}\textbackslash{n}\textbackslash{n})} & \textbf{71.7 {\scriptsize $\pm$0.26}} &  \textbf{49.3 {\scriptsize $\pm$0.19}} & \textbf{77.4 {\scriptsize $\pm$0.16}} \\
        \multicolumn{2}{l}{\ours{} (\texttt{TripleHash}, \ie \#\#\#)} & 70.6 {\scriptsize $\pm$0.09} & 47.1 {\scriptsize $\pm$0.33} & 76.9 {\scriptsize $\pm$0.08} \\
        \multicolumn{2}{l}{\ours{} (\texttt{TripleStar}, \ie ***)} & 70.9 {\scriptsize $\pm$0.14} & 46.3 {\scriptsize $\pm$0.78} & 76.7 {\scriptsize $\pm$0.24} \\
        \multicolumn{2}{l}{\ours{} (\texttt{<br>})} & 71.6 {\scriptsize $\pm$0.29} & 46.6 {\scriptsize $\pm$1.10} & 76.9 {\scriptsize $\pm$0.08} \\
        \multicolumn{2}{l}{\ours{} (\texttt{<br/>})} & 70.0 {\scriptsize $\pm$0.43} & 45.8 {\scriptsize $\pm$0.94} & 76.5 {\scriptsize $\pm$0.24} \\
        \midrule
        \multicolumn{2}{l}{Heterogeneous \ours{}} & 71.3 {\scriptsize $\pm$0.42} & 47.6 {\scriptsize $\pm$0.52} & 76.9 {\scriptsize $\pm$0.14} \\
        \bottomrule
    \end{tabular}
    \label{table1}
\end{table*}
We report the statistics of accuracy values obtained by running the experiments for 3 trials.

\subsection{Results}

\noindent Table \ref{table1} compares the performances of vanilla CoT and \ours{} tested on three reasoning tasks, for GPT-3.5-Turbo. One can confirm that LLMs attain higher performances with the inclusion of separators (specifically, \textbackslash{n}\textbackslash{n}\textbackslash{n}) at the end of each prompt exemplar. This validates that the insertion of separators after each exemplar significantly improves LLM reasoning. \\
\\
We also test \ours{} on other models including Llama2-7b, GPT-4-turbo and GPT-4, in  Table~\ref{table2}. It turns out that compared with vanilla CoT, adding separators (either \texttt{TripleSkip} or \texttt{TripleHash}) increases the accuracy on most cases. For example, in GPT-4-turbo model, \ours{} (\texttt{TripleSkip}) enjoys 5.1\% and 2.6\% gap for GSM8K and AQuA datasets, respectively.\\
\\
Now, the follow-up questions are: \\
\subsubsection{Which Separator is Effective in Improving Reasoning?} We test 5 different separators: (1)  \texttt{TripleSkip} (\textbackslash{n}\textbackslash{n}\textbackslash{n}) which uses three newline separators, (2) \texttt{TripleHash} (\#\#\#), (3) \texttt{TripleStar} (***), and two versions of HTML line break tags: (4) \verb|<br>| and (5) \verb|<br/>|. 
Table \ref{table1} shows that although the use of various separators in \ours{} improves LLM reasoning compared to the vanilla CoT, \texttt{TripleSkip} is the most effective separator for enhancing LLM reasoning for the tested datasets, on GPT-3.5-Turbo. It is true that some separators are hurting the accuracy (\eg inserting \verb|<br/>| separator after exemplars is not desirable for GSM8K and AQuA datasets), which implies that it is necessary to appropriately choose separators. \\
\begin{table*}[tp]
\scriptsize
\setlength\tabcolsep{3pt}
    \caption{Comparison of Vanilla CoT and \ours{}  on various LLMs including Llama2-7b, GPT-4-turbo and GPT-4. Here, the number of maximum generated tokens is set to 128. In most cases, \ours{} outperforms vanilla CoT. \\
    {\footnotesize (It should be noted that for the Llama2-7b model, all three trials produced identical accuracy results. Consequently, the standard deviation is not reported.)}}
    \centering
    \vspace{0.4cm}
    \label{experimental-result}
    \begin{tabular}{@{}lcccccccccc@{}}
        \toprule
        
        & & \multicolumn{3}{c}{Llama2-7b} & \multicolumn{3}{c}{GPT-4-turbo} & \multicolumn{3}{c}{GPT-4}\\ \cmidrule(lr){3-5} \cmidrule(lr){6-8} \cmidrule{9-11}
        Method& & \multicolumn{1}{c}{GSM8K} & \multicolumn{1}{c}{AQuA} & \multicolumn{1}{c}{CSQA}&
        \multicolumn{1}{c}{GSM8K} &\multicolumn{1}{c}{AQuA} & \multicolumn{1}{c}{CSQA} & \multicolumn{1}{c}{GSM8K} &\multicolumn{1}{c}{AQuA} & \multicolumn{1}{c}{CSQA}\\

        \midrule 

        \multicolumn{2}{l}{Vanilla CoT} & 14.7 & 19.3 &62.2 & 63.0 {\scriptsize $\pm$ 0.95} & 31.4 {\scriptsize $\pm$ 1.32} & \textbf{86.4} {\scriptsize $\pm$0.31} & 89.4 {\scriptsize $\pm$ 0.19} & 71.1 {\scriptsize $\pm$ 0.78} & 86.4 {\scriptsize $\pm$ 0.14} \\
        \multicolumn{2}{l}{\ours{} (\texttt{TripleSkip})} & 13.4 &  19.3 & 62.4 &  \textbf{68.1} {\scriptsize $\pm$ 0.45} & \textbf{34.0} {\scriptsize $\pm$ 0.65} & 86.0 {\scriptsize $\pm$0.65} &88.6 {\scriptsize $\pm$ 0.29} & \textbf{71.5} {\scriptsize $\pm$ 0.82} & \textbf{86.5} {\scriptsize $\pm$ 0.24} \\
        \multicolumn{2}{l}{\ours{} (\texttt{TripleHash})} & \textbf{15.1} &  19.3 & \textbf{62.7} &  67.5 {\scriptsize $\pm$ 0.19} & 34.0 {\scriptsize $\pm$1.90} & 86.1 {\scriptsize $\pm$ 0.22} & \textbf{89.6} {\scriptsize $\pm$ 0.12} & 69.6 {\scriptsize $\pm$ 1.47} & 86.3 {\scriptsize $\pm$ 0.22} \\
        \bottomrule
    \end{tabular}
    \label{table2}
\end{table*} 
\\
In a situation when there is uncertainty about which separator is the appropriate choice in a specific task, one might wonder which separator should be chosen. To answer this question, we test with a variant of our method, dubbed as Heterogeneous \ours{}, which uses different separators for different exemplars within the prompt. Recall that we consider 5 different types of separators, \texttt{TripleSkip}, \texttt{TripleHash}, \texttt{TripleStar},
\verb|<br>| and \verb|<br/>|. These types of separators are placed in turn after distinct exemplars. See Table~\ref{table:gsm8k_hetero} for an example prompt for Heterogeneous \ours{}. Our result in Table~\ref{table1} shows that Heterogeneous \ours{} outperforms vanilla CoT, implying that even in the absence of a clear choice for the best separator, \ours{} improves LLM reasoning compared to vanilla CoT. Thus, Heterogeneous \ours{} can be used in practical scenarios where we do not have any prior knowledge on the optimal separator for our new target task.\\

\begin{table*}[h!]
\centering
\small
\caption{Full Prompt for Heterogeneous \ours{} used in our experiments on the GSM8K Arithmetic Reasoning Benchmark. More prompt exemplars are available in Appendix. \ref{prompt_exemplars}}
\vspace{0.3cm}
\begin{tabular}{p{13cm}}
\hline
\vspace{0.1cm}
Q: There are 15 trees in the grove. Grove workers will plant trees in the grove today. After they are done, there will be 21 trees. How many trees did the grove workers plant today? \\
A: There are 15 trees originally. Then there were 21 trees after some more were planted. So there must have been 21 - 15 = 6. The answer is 6. \\
\colorbox{pastelblue}{\textbackslash{n}} \\
\colorbox{pastelblue}{\textbackslash{n}} \\
\colorbox{pastelblue}{\textbackslash{n}} \\
Q: If there are 3 cars in the parking lot and 2 more cars arrive, how many cars are in the parking lot? \\
A: There are originally 3 cars. 2 more cars arrive. 3 + 2 = 5. The answer is 5. \colorbox{pastelblue}{\#\#\#} \\
Q: Leah had 32 chocolates and her sister had 42. If they ate 35, how many pieces do they have left in total? \\
A: Originally, Leah had 32 chocolates. Her sister had 42. So in total they had 32 + 42 = 74. After eating 35, they
had 74 - 35 = 39. The answer is 39. \colorbox{pastelblue}{***} \\
Q: Jason had 20 lollipops. He gave Denny some lollipops. Now Jason has 12 lollipops. How many lollipops did Jason give to Denny? \\
A: Jason started with 20 lollipops. Then he had 12 after giving some to Denny. So he gave Denny 20 - 12 = 8. The answer is 8. \colorbox{pastelblue}{\texttt{<br>}}\\
Q: Shawn has five toys. For Christmas, he got two toys each from his mom and dad. How many toys does he have now? \\
A: Shawn started with 5 toys. If he got 2 toys each from his mom and dad, then that is 4 more toys. 5 + 4 = 9. The answer is 9. \colorbox{pastelblue}{\texttt{<br/>}} \\
Q: There were nine computers in the server room. Five more computers were installed each day, from monday to thursday. How many computers are now in the server room? \\
A: There were originally 9 computers. For each of 4 days, 5 more computers were added. So 5 * 4 = 20 computers were added. 9 + 20 is 29. The answer is 29. \\
\colorbox{pastelblue}{\textbackslash{n}} \\
\colorbox{pastelblue}{\textbackslash{n}} \\
\colorbox{pastelblue}{\textbackslash{n}} \\
Q: Michael had 58 golf balls. On tuesday, he lost 23 golf balls. On wednesday, he lost 2 more. How many golf balls did he have at the end of wednesday? \\
A: Michael started with 58 golf balls. After losing 23 on tuesday, he had 58 - 23 = 35. After losing 2 more, he had 35 - 2 = 33 golf balls. The answer is 33. \colorbox{pastelblue}{\#\#\#} \\
Q: Olivia has \$23. She bought five bagels for \$3 each. How much money does she have left? \\
A: Olivia had 23 dollars. 5 bagels for 3 dollars each will be 5 x 3 = 15 dollars. So she has 23 - 15 dollars left. 23 - 15 is 8. The answer is 8. \colorbox{pastelblue}{***}
\vspace{0.1cm}
\\
\hline
\end{tabular}
\label{table:gsm8k_hetero}
\end{table*}

\subsubsection{Does the Location/Number of Separators used in \ours{} Contribute to Enhancing CoT prompting?}  
\begin{figure*}[ht]
\begin{center}
\includegraphics[scale=0.25]{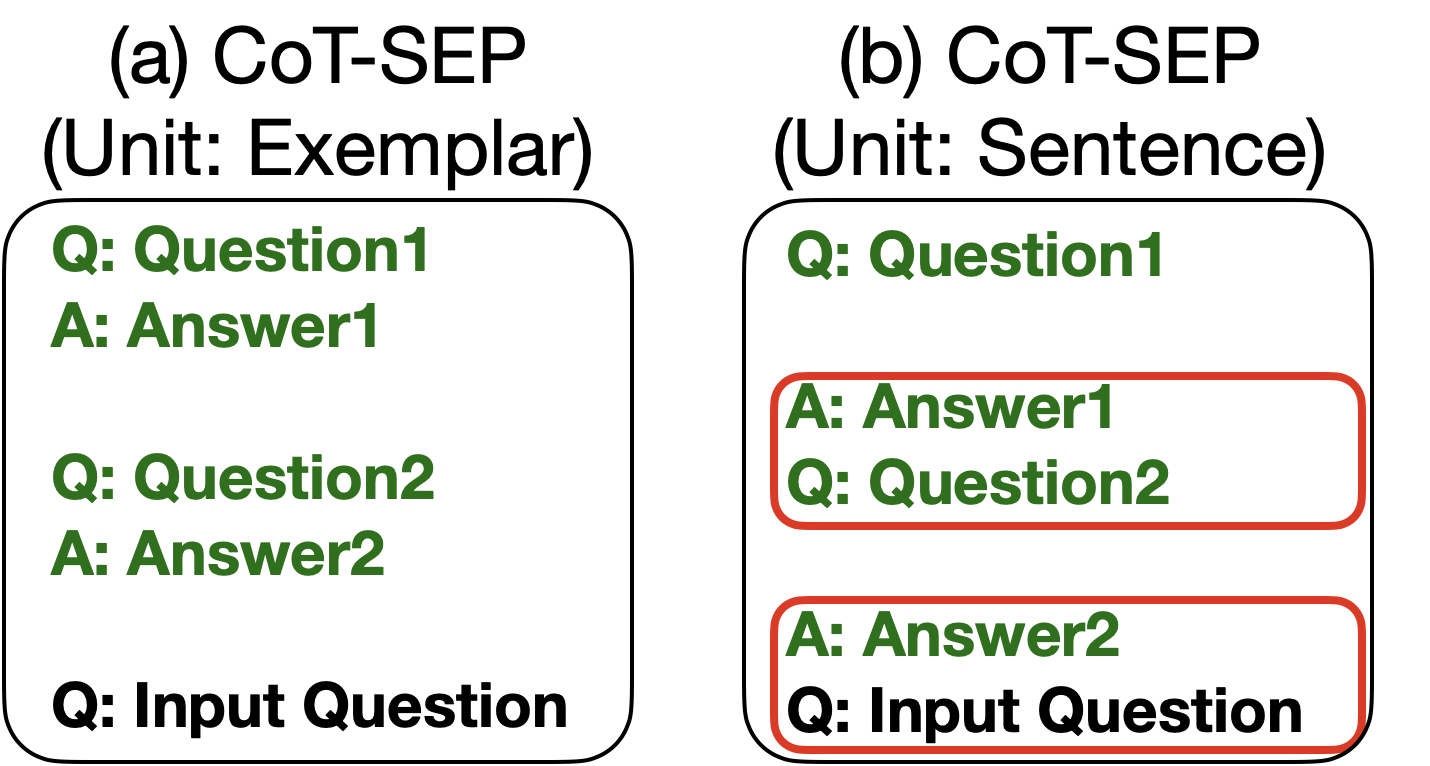}
\end{center}
\caption{
The prompts used for COT-SEP (Unit: Exemplar) and COT-SEP (Unit: Sentence). The former one
is more interpretable, which provides a better performance in Table~\ref{table:ablation}. See Fig.~\ref{fig:figure2-lengthy} for a detailed figure with
exemplars.}
\label{fig:figure2}
\end{figure*}

To check the effect of the location of separators, 
we experiment \ours{} with two versions, \ours{} (Unit: Exemplar) and \ours{} (Unit: Sentence), as shown in Fig.~\ref{fig:figure2}. The first version adds the separator at the end of each exemplar, while the second version adds the separator between sentences within each exemplar's CoT. 
Table~\ref{table:ablation} shows that \ours{} (Unit: Exemplar) outperforms \ours{} (Unit: Sentence). This can be explained by the visualization in Fig.~\ref{fig:figure2}: 
in the prompt of \ours{} (Unit: Sentence), 
the next question (say Question2) and the answer of the previous question (say Answer1) appear as a set, making it difficult for not only LLMs but also humans to understand, which results in the performance worse than \ours{} and even vanilla CoT. This shows that the placement of separators is crucial.
Table~\ref{table:ablation}  also shows the effect of the number of separators.
The result indicates that our \ours{} (\texttt{TripleSkip}) method, with its specific number of three newline separators, achieves the best performance in GSM8K and AQuA benchmarks compared to the reduced number of separators. \ours{} (\texttt{TripleSkip}) is also on par with the highest performance for the CSQA benchmark. \\

\subsubsection{Why does the Performance Gain of \ours{} differ by Reasoning Tasks?}
As illustrated in Tables \ref{table1}, \ref{table2}, \ref{table:ablation}, we can observe that the performance gain of \ours{}, regardless of which separator is being used, varies for different reasoning tasks. We hypothesize the main reason for this variance is due to the variance of performance of CoT itself on different datasets. In particular, based on the explanation provided by \cite{wei2022chain}, we identify \ours{} to have the most gain when the task in question is challenging. Tables \ref{table1}, \ref{table2}, and \ref{table:ablation} clearly show that the AQuA reasoning benchmark has the lowest overall accuracy, indicating that it is the most challenging. Therefore, \ours{} demonstrates a greater performance improvement on the AQuA benchmark compared to the other benchmarks.

\begin{table*}[ht]
\caption{The effect of location/number of separators in \ours{} for GPT-3.5-Turbo model, when the separator is `\textbackslash{n}'. 
We consider two scenarios having different \textit{units} of applying separators: the first scenario adds separators at the end of each exemplar, while the second one adds separators between sentences within each exemplar's CoT. Note that adding the separator 3 times in the first scenario reduces to \ours{}(\texttt{TripleSkip}) in Table~\ref{table1}. Adding separators after each exemplars has up to 3.5\% accuracy gain, compared with adding separators between sentences. This result is supported by our visualization in Fig.~\ref{fig:figure2}.   
}
\vspace{0.3mm}
\centering
\begin{tabular}{llcccc}
\toprule
& \multicolumn{2}{c}{Arithmetic Reasoning} & \multicolumn{1}{l}{Commonsense Reasoning} \\ \cmidrule(lr){2-3} \cmidrule(lr){4-4}   
Method & \multicolumn{1}{c}{GSM8K} & \multicolumn{1}{c}{AQuA} & \multicolumn{1}{c}{CSQA} \\ 
\midrule 
        Vanilla CoT & 70.4 {\scriptsize $\pm$0.17} & 46.5 {\scriptsize $\pm$0.82} & 76.5 {\scriptsize $\pm$0.14} \\
\midrule
\underline{\ours{} (Unit : Exemplar)} \\
\quad \textbackslash{n} & 71.0 {\scriptsize $\pm$0.29} & 46.0 {\scriptsize $\pm$0.68} &\textbf{77.5 {\scriptsize $\pm$0.12}}\\
\quad \textbackslash{n}\textbackslash{n} & 70.5 {\scriptsize $\pm$0.22} & 47.6 {\scriptsize $\pm$0.83} & 77.0 {\scriptsize $\pm$0.12} \\
\quad \textbackslash{n}\textbackslash{n}\textbackslash{n} (\texttt{TripleSkip}) & \textbf{71.7 {\scriptsize $\pm$0.26}} & \textbf{49.3 {\scriptsize $\pm$0.19}} & 77.4 {\scriptsize $\pm$0.16} \\
\midrule
\underline{\ours{} (Unit : Sentence)} \\
\quad \textbackslash{n} & 69.9 {\scriptsize $\pm$0.26} & 44.5 {\scriptsize $\pm$0.33} & \textbf{75.9 {\scriptsize $\pm$0.12}} \\
\quad \textbackslash{n}\textbackslash{n} & 70.2 {\scriptsize $\pm$0.22} & 44.2 {\scriptsize $\pm$0.19} & 74.5 {\scriptsize $\pm$0.12} \\
\quad \textbackslash{n}\textbackslash{n}\textbackslash{n} & \textbf{70.8 {\scriptsize $\pm$0.17}} & \textbf{45.8 {\scriptsize $\pm$1.00}} & 75.2 {\scriptsize $\pm$0.22} \\
\bottomrule
\end{tabular}
\label{table:ablation}
\end{table*}

\section{Limitations}
\noindent Our research relies on CoT prompting \cite{wei2022chain}, which is based on in-context learning. Given CoT prompting, the model outputs a final answer along with a step-by-step reasoning process, allowing the generated CoT to validate the final answer. This process may cause users to excessively trust LLMs, which is critical, as LLMs may often generate incorrect outputs. This may ultimately lead to automation bias. Therefore, users should be aware of the need not to overly rely on LLMs for solving reasoning tasks or to make all decisions, despite receiving aid from them. \\
\\
For our \ours{} framework, we utilize various separators, \texttt{TripleSkip} (\textbackslash{n}\textbackslash{n}\textbackslash{n}), \texttt{TripleHash} (\#\#\#), \texttt{TripleStar} (***), \texttt{<br>}, and \texttt{<br/>}. However, as shown in our various experiments, the results of the LLMs depend greatly on where and how the separators are located within the prompts. Therefore, it is recommended to cautiously place separators in the precise location stated in our framework, otherwise, the accuracy of the outcome may be diminished. \\   
\\
Moreover, due to the expense of large language models (LLMs), this study omits larger datasets and additional task categories, which may limit further insights. Furthermore, due to the limited access to computational resources, we conducted a single trial using Meta's \texttt{Llama-2-7b} model, while carrying out three trials with OpenAI's GPT models. For the same reason, we only focus on a few datasets to validate our method. We tested on two arithmetic reasoning datasets, GSM8K \cite{cobbe2021training} and AQuA \cite{ling2017program}, as they are the most challenging arithmetic reasoning benchmarks available, and had the least impressive performance while using vanilla CoT prompting. For validation on another reasoning task, we selected the CSQA \cite{talmor-etal-2019-commonsenseqa} commonsense reasoning benchmark, as it is one of the most difficult commonsense reasoning benchmarks. 

\section{Conclusion}
\noindent In this paper, we introduce \ours{}, a method placing separators in a structured format to improve Chain-of-Thought (CoT) prompting popular in large language models. Within this framework, we place separators at the end of each prompt exemplar for better readability for LLMs. We demonstrate that \ours{} effectively improves CoT prompting in various LLMs across highly complex arithmetic and commonsense benchmarks through multiple experiments employing different separators and structural variations. While our experimental results are only focused on the CoT prompting scenarios, our simple idea of adding separators within the prompt can be easily plugged into various existing prompting techniques, such as Generated Knowledge Prompting \cite{liu2021generated}, Self-Consistency \cite{wang2023selfconsistency}, Tree of Thought \cite{yao2023tree, long2023llmtot} and GraphPrompt \cite{10.1145/3543507.3583386}. We look forward to applying our approach to different prompting strategies and various LLMs in a near future.

\bibliographystyle{IEEEtran}
\bibliography{custom}

\newpage
\onecolumn
\appendices
\section{Experiment Details for Difference in Separator Location in \ours{}}
\label{sec:difference_csep}
\begin{figure*}[hb]
\begin{center}
\includegraphics[scale=0.20]{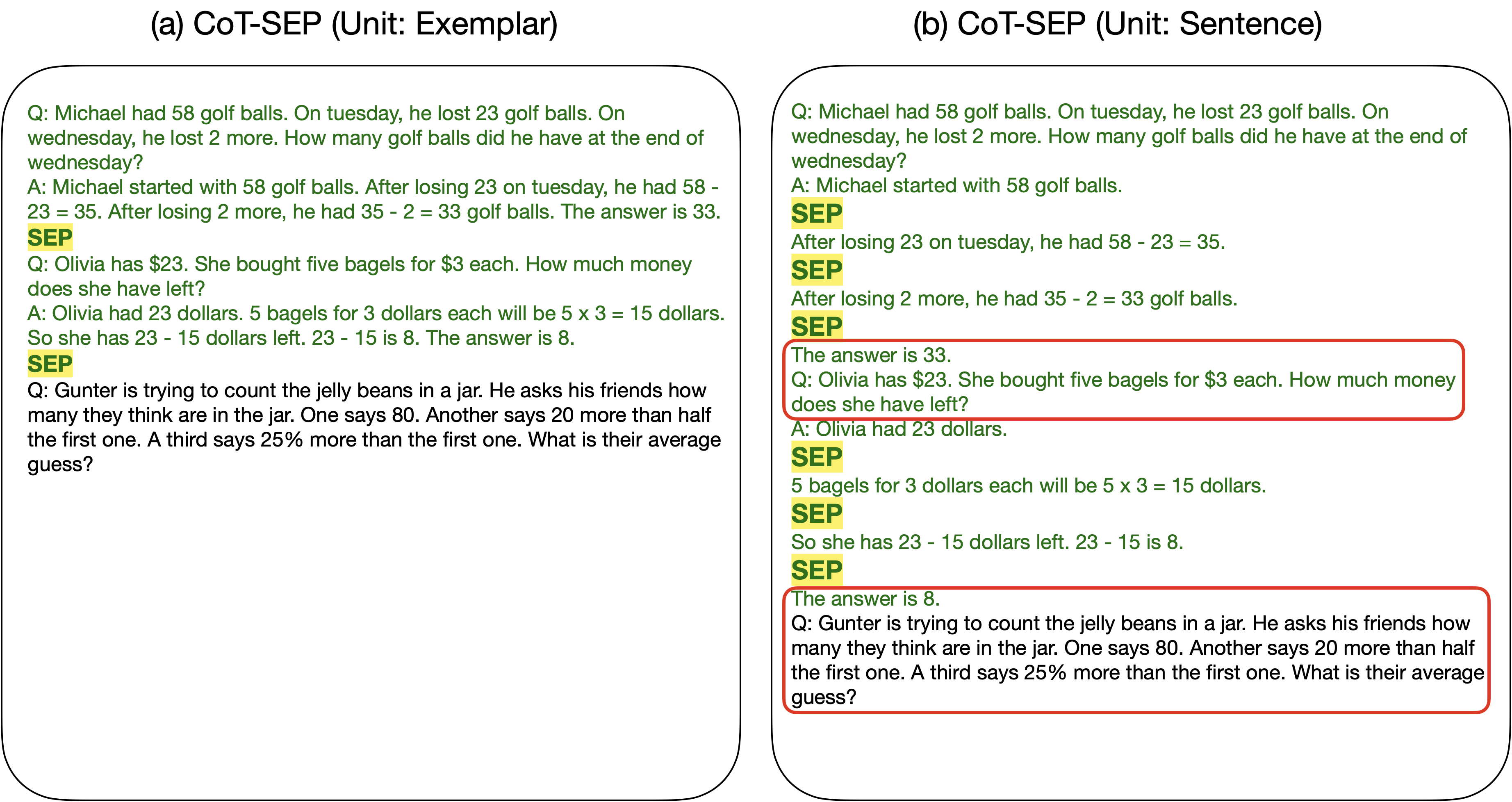}
\end{center}
\caption{The prompts used for \ours{} (Unit: Exemplar) and \ours{} (Unit: Sentence) on GSM8K with two exemplars.
In the first version, the separator (denoted as SEP in the above figure) is inserted after each exemplar, while in the second version, the separator is added between consecutive sentences of the Chain-of-Thought. 
The former one is more interpretable, which provides a better performance in Table~\ref{table:ablation}.}
\label{fig:figure2-lengthy}
\end{figure*}

In this section, we explain our experiment shown in Table~\ref{table:ablation} in more detail. We provide Fig.~\ref{fig:figure2-lengthy}, which is a detailed figure of Fig.~\ref{fig:figure2} with two exemplars.

\section{Full List of Prompt Exemplars} \label{prompt_exemplars} 
Our prompts are mainly derived from the standard Chain-of-Thought (CoT) prompts provided by Wei et al. (2022). The most notable prompts used by our method, \ours{}, are included in the appendix. Complete prompts for our experimental setup can be found at \href{https://github.com/cottonlove/CoT-SEP}{\texttt{https://github.com/cottonlove/CoT-SEP}}.

\vspace{-0.1cm}
\begin{table*}[h!]
\centering
\scriptsize
\setlength{\tabcolsep}{2pt}
\caption{Full prompt for \ours{} (\texttt{TripleSkip}) used in our experiments on the AQuA Arithmetic Reasoning Benchmark.}
\vspace{0.3cm}
\begin{tabular}{p{15cm}}
\hline
\vspace{0.1cm}
Q: John found that the average of 15 numbers is 40. If 10 is added to each number then the mean of the numbers \\ is? Answer Choices: (a) 50 (b) 45 (c) 65 (d) 78 (e) 64 \\
A: If 10 is added to each number, then the mean of the numbers also increases by 10. So the new mean would be 50. The answer is (a). \\
\colorbox{pastelblue}{\textbackslash{n}}\\
\colorbox{pastelblue}{\textbackslash{n}}\\
\colorbox{pastelblue}{\textbackslash{n}}\\
Q: If a / b = 3/4 and 8a + 5b = 22,then find the value of a. Answer Choices: (a) 1/2 (b) 3/2 (c) 5/2 (d) 4/2 (e) 7/2 \\
A: If a / b = 3/4, then b = 4a / 3. So 8a + 5(4a / 3) = 22. This simplifies to 8a + 20a / 3 = 22, which means 44a / 3 = 22. So a is equal to 3/2. The answer is (b). \\
\colorbox{pastelblue}{\textbackslash{n}}\\
\colorbox{pastelblue}{\textbackslash{n}}\\
\colorbox{pastelblue}{\textbackslash{n}}\\
Q: A person is traveling at 20 km/hr and reached his destiny in 2.5 hr then find the distance? Answer Choices: (a) 53 km (b) 55 km (c) 52 km (d) 60 km (e) 50 km \\
A: The distance that the person traveled would have been 20 km/hr * 2.5 hrs = 50 km. The answer is (e). \\
\colorbox{pastelblue}{\textbackslash{n}}\\
\colorbox{pastelblue}{\textbackslash{n}}\\
\colorbox{pastelblue}{\textbackslash{n}}\\
Q: How many keystrokes are needed to type the numbers from 1 to 500? Answer Choices: (a) 1156 (b) 1392 (c) 1480 (d) 1562 (e) 1788 \\
A: There are 9 one-digit numbers from 1 to 9. There are 90 two-digit numbers from 10 to 99. There are 401 three-digit numbers from 100 to 500. 9 + 90(2) + 401(3) = 1392. The answer is (b).\\
\colorbox{pastelblue}{\textbackslash{n}}\\
\colorbox{pastelblue}{\textbackslash{n}}\\
\colorbox{pastelblue}{\textbackslash{n}}
\vspace{0.1cm}
\\
\hline
\end{tabular}
\label{table:aqua_tripleskip}
\end{table*}

\begin{table*}[h!]
\centering
\scriptsize
\caption{Full prompt for \ours{} (\texttt{TripleSkip}) used in our experiments on the CSQA Commonsense Reasoning Benchmark.}
\vspace{0.3cm}
\begin{tabular}{p{15cm}}
\hline
\vspace{0.1cm}
Q: What do people use to absorb extra ink from a fountain pen? Answer Choices: 
(a) shirt pocket 
(b) calligrapher’s hand 
(c) inkwell 
(d) desk drawer 
(e) blotter \\
A: The answer must be an item that can absorb ink. Of the above choices, only blotters are used to absorb ink. So the answer is (e). \\
\colorbox{pastelblue}{\textbackslash{n}}\\
\colorbox{pastelblue}{\textbackslash{n}}\\
\colorbox{pastelblue}{\textbackslash{n}}\\
Q: What home entertainment equipment requires cable? Answer Choices: 
(a) radio shack 
(b) substation 
(c) television 
(d) cabinet\\
A: The answer must require cable. Of the above choices, only television requires cable. So the answer is (c). \\
Q: The fox walked from the city into the forest, what was it looking for? Answer Choices: 
(a) pretty flowers 
(b) hen house 
(c) natural habitat 
(d) storybook \\
A: The answer must be something in the forest. Of the above choices, only natural habitat is in the forest. So the answer is (b). \\
\colorbox{pastelblue}{\textbackslash{n}}\\
\colorbox{pastelblue}{\textbackslash{n}}\\
\colorbox{pastelblue}{\textbackslash{n}}\\
Q: Sammy wanted to go to where the people were. Where might he go? Answer Choices: 
(a) populated areas 
(b) race track 
(c) desert 
(d) apartment 
(e) roadblock\\
A: The answer must be a place with a lot of people. Of the above choices, only populated areas have a lot of people. So the answer is (a).\\
\colorbox{pastelblue}{\textbackslash{n}}\\
\colorbox{pastelblue}{\textbackslash{n}}\\
\colorbox{pastelblue}{\textbackslash{n}}\\
Q: Where do you put your grapes just before checking out? Answer Choices: 
(a) mouth 
(b) grocery cart 
(c)super market 
(d) fruit basket 
(e) fruit market\\
A: The answer should be the place where grocery items are placed before checking out. Of the above choices, grocery cart makes the most sense for holding grocery items. So the answer is (b).\\
\colorbox{pastelblue}{\textbackslash{n}}\\
\colorbox{pastelblue}{\textbackslash{n}}\\
\colorbox{pastelblue}{\textbackslash{n}}\\
Q: Google Maps and other highway and street GPS services have replaced what? Answer Choices: 
(a) united states 
(b) mexico 
(c) countryside 
(d) atlas\\
A: The answer must be something that used to do what Google Maps and GPS services do, which is to give directions. Of the above choices, only atlases are used to give directions. So the answer is (d).\\
\colorbox{pastelblue}{\textbackslash{n}}\\
\colorbox{pastelblue}{\textbackslash{n}}\\
\colorbox{pastelblue}{\textbackslash{n}}\\
Q: Before getting a divorce, what did the wife feel who was doing all the work? Answer Choices: 
(a) harder 
(b) anguish 
(c) bitterness 
(d) tears 
(e) sadness\\
A: The answer should be the feeling of someone getting divorced who was doing all the work. Of the above choices, the closest feeling is bitterness. So the answer is (c). \\
\colorbox{pastelblue}{\textbackslash{n}}\\
\colorbox{pastelblue}{\textbackslash{n}}\\
\colorbox{pastelblue}{\textbackslash{n}}
\vspace{0.1cm}
\\
\hline
\end{tabular}
\label{table:csqa_tripleskip}
\end{table*}

\end{document}